\documentclass[12pt]{elsarticle}

\usepackage{amssymb}
\usepackage{amsmath}
\usepackage{graphicx}
\usepackage[colorlinks=true, allcolors=blue]{hyperref}
\usepackage{multirow}
\newcommand{\ts}[1]{\textsubscript{#1}}

\usepackage{wrapfig}
\usepackage{booktabs}
\usepackage{appendix}

\usepackage[symbol]{footmisc}
\usepackage{algorithm} % Required for pseudo code
\usepackage[noend]{algpseudocode} % Required for pseudo code

\newcommand\NoDo{\renewcommand\algorithmicdo{}}

\usepackage[capitalize]{cleveref}
\crefname{section}{Sec.}{Secs.}
\Crefname{section}{Section}{Sections}
\Crefname{table}{Table}{Tables}
\crefname{table}{Tab.}{Tabs.}

\journal{Pattern Recognition Letters}

\begin{document}

\begin{frontmatter}

%% Title, authors and addresses

%% use the tnoteref command within \title for footnotes;
%% use the tnotetext command for theassociated footnote;
%% use the fnref command within \author or \address for footnotes;
%% use the fntext command for theassociated footnote;
%% use the corref command within \author for corresponding author footnotes;
%% use the cortext command for theassociated footnote;
%% use the ead command for the email address,
%% and the form \ead[url] for the home page:
%% \title{Title\tnoteref{label1}}
%% \tnotetext[label1]{}
%% \author{Name\corref{cor1}\fnref{label2}}
%% \ead{email address}
%% \ead[url]{home page}
%% \fntext[label2]{}
%% \cortext[cor1]{}
%% \affiliation{organization={},
%%             addressline={},
%%             city={},
%%             postcode={},
%%             state={},
%%             country={}}
%% \fntext[label3]{}

\title{Feature Perturbation Augmentation for Reliable Evaluation of Importance Estimators in Neural Networks\footnote[1]{Early version at the ICLR 2023 Workshop on Trustworthy ML; Full paper published in \emph{Pattern Recognition Letters} \href{https://www.sciencedirect.com/science/article/pii/S0167865523002842}{doi:10.1016/j.patrec.2023.10.012}}}

%% use optional labels to link authors explicitly to addresses:
%% \author[label1,label2]{}
%% \affiliation[label1]{organization={},
%%             addressline={},
%%             city={},
%%             postcode={},
%%             state={},
%%             country={}}
%%
%% \affiliation[label2]{organization={},
%%             addressline={},
%%             city={},
%%             postcode={},
%%             state={},
%%             country={}}

\author[]{Lennart Brocki\corref{email: brocki.lennart@gmail.com}}
\author[]{Neo Christopher Chung\corref{email: ncnchung@gmail.com}}

\affiliation[]{organization={Institute of Informatics, University of Warsaw},%Department and Organization
            addressline={Banacha 2}, 
            city={Warsaw},
            postcode={02-097}, 
            state={},
            country={Poland}
            }
            
\begin{abstract}
Post-hoc explanation methods attempt to make the inner workings of deep neural networks more comprehensible and trustworthy, which otherwise act as black box models. However, since a ground truth is in general lacking, local post-hoc explanation methods, which assign importance scores to input features, are challenging to evaluate. One of the most popular evaluation frameworks is to perturb features deemed important by an explanation and to measure the change in prediction accuracy. Intuitively, a large decrease in prediction accuracy would indicate that the explanation has correctly quantified the importance of features with respect to the prediction outcome (e.g., logits). However, the change in the prediction outcome may stem from perturbation artifacts, since perturbed samples in the test dataset are out of distribution (OOD) compared to the training dataset and can therefore potentially disturb the model in an unexpected manner. To overcome this challenge, we propose feature perturbation augmentation (FPA) which creates and adds perturbed images during the model training. Our computational experiments suggest that FPA makes the considered models more robust against perturbations. Overall, FPA is an intuitive and straightforward data augmentation technique that renders the evaluation of post-hoc explanations more trustworthy \newline 

\noindent Reproducible codes and pre-trained models with FPA are available on Github: \url{https://github.com/lenbrocki/Feature-Perturbation-Augmentation}.
\end{abstract}

% %%Graphical abstract
% \begin{graphicalabstract}
% \includegraphics{grabs}
% \end{graphicalabstract}

%%Research highlights
%\begin{highlights}
%\item Post-hoc explanation methods often measure the accuracy decrease after feature perturbation.
%\item The accuracy decrease may stem from perturbation artifacts, instead of information removal. 
%\item Feature Perturbation Augmentation removes perturbation artifacts when explaining DNNs.
%\item Perturbation curves show that FPA enables more reliable evaluation of importance estimators.
%\end{highlights}

\begin{keyword}
%% keywords here, in the form: keyword \sep keyword
deep neural network \sep interpretability \sep explainability \sep importance estimator \sep saliency map \sep data augmentation 
% %% PACS codes here, in the form: \PACS code \sep code
% \PACS 0000 \sep 1111
% %% MSC codes here, in the form: \MSC code \sep code
% %% or \MSC[2008] code \sep code (2000 is the default)
% \MSC 0000 \sep 1111
\end{keyword}

\end{frontmatter}

%% \linenumbers

\section{Introduction}
Deep learning exhibits state-of-the-art performance in a wide range of computer vision tasks. However, the reasons underlying classifications and predictions made by deep neural networks (DNN) are difficult to extract due to their nested non-linear structure and a large number of parameters \citep{samek2019explainable}. Post-hoc explanations, which estimate the importance of input features with respect to the model's output \citep{simonyan2014deep, smilkov2017smoothgrad, sundararajan2017axiomatic}, are often used to make deep learning models more interpretable. However, evaluating the fidelity of post-hoc importance estimators is highly convoluted due to a lack of ground truth and the issue of unintentionally triggering perturbation artifacts. In this study, we introduce \emph{feature perturbation augmentation (FPA)} which aims to avoid the pitfalls of a perturbation-based evaluation of interpretability methods.

%An open question is whether the explanations produced by such importance estimators can be trusted to reliably describe a model's reasoning \cite{adebayo2018sanity, rudin2019stop}.

% \begin{wrapfigure}{R}{0.4\textwidth}
% \centering
% \includegraphics[width=0.4\textwidth]{graphics/cifar_intgrad-5.svg}
% \caption{Perturbation curves for models trained with different data augmentations. Importance scores are obtained using integrated gradients on CIFAR-10 dataset. Top: Most important pixels are masked first (MIF). Bottom: Least important pixels first (LIF).} 
% \label{fig1}
% \end{wrapfigure}

A promising approach for comparing importance estimators despite the aforementioned lack of ground truth is the perturbation of input features \citep{samek2016evaluating,petsiuk2018rise,kindermans2017learning}. Conceptually, if the model's accuracy rapidly decreases by masking pixels deemed most important by some estimator, then it can be concluded that the considered estimator describes the model more accurately than others which result in a slower decrease. However, such an evaluation by perturbation may be problematic due to the risk of unwittingly triggering artifacts of the deep learning model \citep{hooker2019benchmark,fong2017interpretable}. In other words, even when truly unimportant pixels are masked, the accuracy might decrease considerably nonetheless, casting doubt on the reliability of the perturbation-based evaluation approach. This issue is closely related to adversarial examples \cite{goodfellow2014explaining}, which are small, but worst-case, perturbations of input images that cause the model to output a wrong prediction with high confidence. 

Our proposed approach mitigates the influence of perturbation artifacts by training the model with data augmentation that reflects the perturbation used in the evaluation frameworks. We apply the proposed methods on three datasets (CIFAR-10 \cite{cifar}, Food101 \cite{bossard2014food}, the ImageNet \cite{deng2009imagenet}), using four different post-hoc explanation methods. Subsequently, we measure the model output while perturbing an increasing fraction of input features sorted either in most important first (MIF) or least important first (LIF) order. MIF and LIF perturbation curves demonstrate that when using FPA during training, the resulting model exhibits increased robustness against perturbation artifacts, and the evaluation of importance estimators becomes more reliable.% In particular, when input features that are assigned a negative importance score are masked within a FPA-trained model, the corresponding logit values increase as expected. However, this effect is always suppressed when the model is trained without the proposed FPA. 
%This presents evidence that the influence of artifacts is strongly reduced and as a consequence our approach allows for a more reliable evaluation of the relative fidelity of importance estimators, i.e. whether they faithfully explain the model.

In \cref{related}, we briefly review existing evaluation methods for importance estimators and explain how our approach relates to them. \cref{methods} describes in detail the proposed feature perturbation augmentation, the model architectures and datasets, and the importance estimators under consideration. In \cref{results}, we present the results of our experiments, which demonstrate how FPA makes our evaluations more reliable. We also discuss interesting implications of our results concerning the sign fluctuations that importance estimators often exhibit. Finally, the discussion and conclusion follow in \cref{discussion} and \cref{conclusions}, respectively.
%Finally, our results are discussed and compared with ones from the literature in \cref{discussion} and we conclude in \cref{conclusions}.

\section{Related work}\label{related}

% We aim to improve the evaluation of post-hoc explanations for DNNs by using the proposed \emph{feature perturbation augmentation}. Therefore, we provide a brief overview of existing evaluation methods and data augmentation techniques.

The explainability of deep learning is an active and diverse area of research (see \cite{samek2021explaining} for a recent review). In this work, we focus on post-hoc explanations, which aim to render the inner workings of a previously trained model more transparent and understandable. In contrast, inherently interpretable models are designed from the ground up to provide explanations for their decision-making processes \cite{rudin2022interpretable}. In order to characterize and improve post-hoc interpretability methods, it is imperative to consider what are the desired properties (desiderata) and how to evaluate them. Some of the proposed desiderata include, but are not limited to, faithfulness, localization \cite{saporta2022benchmarking,brocki2022evaluation,zhou2016learning}, sparsity\cite{zhou2016learning}, and fulfillment of certain axiomatic properties \cite{sundararajan2017axiomatic}.
%Many desiderata have been formulated and corresponding evaluations, which allow to compare to what degree different explanations fulfill these desiderata, have been proposed.
 
\textit{Faithfulness} or \textit{fidelity} describes how accurately explanation methods estimate the contribution of input features to the model's predictions. Our study falls into this category, and is related to pixel-flipping \cite{bach2015pixel}, region-perturbation \cite{samek2016evaluating}, and Remove and Retrain (ROAR) \cite{hooker2019benchmark}. All these methods perturb input pixels and measure the resulting change in model performance. Our proposed approach is similar to Remove and Retrain (ROAR)\cite{hooker2019benchmark} in so far as it also requires retraining the model but it differs in important aspects. In ROAR, for example, the model is re-trained repeatedly with an increasing fraction of the most important input pixels masked, where the ranking of pixels is established by applying an importance estimator on the original model (trained without any perturbation).
In contrast, the proposed \emph{Feature Perturbation Augmentation} trains the model only once without using the knowledge of the importance estimator and uses the same model for subsequent evaluation.

Other methods to evaluate faithfulness include \textit{faithfulness correlation} \cite{bhatt2020evaluating} and \textit{sensitivity-n} \cite{ancona2017towards}, which measure the correlation between the sum of importance scores of masked pixels and the delta in model output. \cite{dabkowski2017real} proposes to crop images to a region deemed important and feed the resized crop back to the model. It is asserted that a good explanation minimizes the area of the crop while maintaining the model's performance. The method of \textit{Performance Information Curves} \cite{kapishnikov2019xrai} applies a bokeh filter to input images and removes the filter patch-wise starting from the most important patches. The resulting softmax activations are then plotted over the ``entropy'' of the perturbed images. There exist several other methods, which are often variations of one of the methods described above: ROAD \cite{rong2022consistent}, IROF \cite{rieger2020irof}, Infidelity \cite{yeh2019fidelity} and Sufficiency \cite{dasgupta2022framework}.

%\textit{Localisation} describes how well explanations are centered around a region of interest (ROI). Unlike perturbation-based methods, localisation evaluation methods do not probe the model directly. Instead, a human derived ground truth — an ROI —  such as organ and pathology segmentations  \cite{saporta2022benchmarking,brocki2022evaluation}, bounding-boxes for objects of interest \cite{zhou2016learning} or human attention maps \cite{selvaraju2017grad} serve as the benchmark. Other desiderata include sensitivity of explanations with respect to \textit{randomizations} of model parameters \cite{adebayo2018sanity} or targeted logits \cite{sixt2020explanations}, \textit{sparseness} \cite{chalasani2020concise} and fulfillment of certain \textit{axiomatic properties} such as completeness \cite{sundararajan2017axiomatic}.

Data augmentation can make machine learning models more robust and generalizable. From noise injection to utilizing DNNs for creating synthetic data, there are many data augmentation techniques (see a review \cite{shorten2019survey}). In deep learning, one may apply geometric and color manipulations, make use of noise and filters, and modify feature space. The closest approach to the proposed FPA method is ``random erasing'' \cite{zhong2020random}. Random erasing augments the data by randomly selecting a rectangle region in an image and replacing its pixels with non-informative values such as white, black, or random RGB values. When training DNNs with random erasing, the resulting model's performance shows increased robustness against occlusion with random rectangles \cite{zhong2020random}. FPA may be seen as a generalization and mixture of noise injection and random erasing where rectangles of varying sizes are probabilistically used to perturb the input data. Of course, the aim of FPA is very different from other data augmentations techniques, since we focus on improving interpretability rather than classifying performance \citep{dziugaite2020tradeoff}.
% \cite{dziugaite2020tradeoff}.

\section{Methods and Materials}\label{methods}
\subsection*{Feature Perturbation Augmentation}
\begin{figure}[ht!]
\centering
\includegraphics[width=0.9\textwidth]{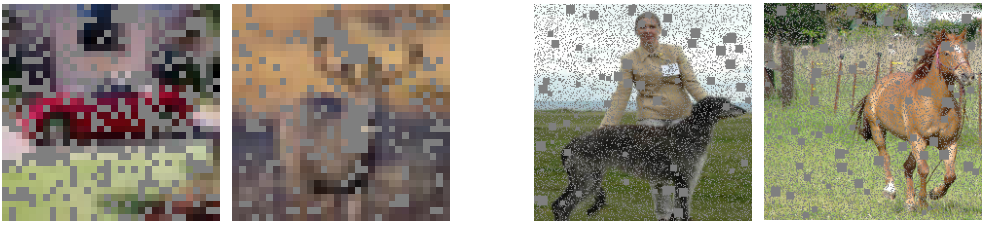}
\caption{Examples of feature perturbation augmentation (FPA) applied on \emph{Left:} CIFAR-10 and \emph{Right:} ImageNet.} 
\label{fig2}
\end{figure}

The perturbation-based evaluation of importance estimators has been criticized \cite{hooker2019benchmark} since the perturbation of input pixels leads to a shift in the data distribution, violating the key assumption that training and test data stem from the same distribution. It is then unclear whether the observed degradation of model performance is due to this out-of-distribution (OOD) problem or the removal of truly informative features. See \cite{fong2017interpretable} for concrete examples of how pixel perturbations can act as adversarial attacks. In fact, \cite{samek2021explaining} argues that certain importance estimators perform very well in perturbation-based evaluations because they strongly trigger perturbation artifacts and not because they faithfully describe the model. 

We propose to overcome this problem by augmenting the data during training using the same kind of data perturbations that will be used for the subsequent evaluation of post-hoc importance estimators. If features are to be masked by setting them to zero, as in our experiments, the training data shall be augmented with samples that have randomly selected pixels set to zero. In this way, FPA mitigates the risk of the OOD problem and one can be more confident that a perturbation-based evaluation of importance estimators actually quantifies the removal of information that is relevant for the model's predictions.

In FPA, mini-batches in training are selected for perturbation with a probability $p$. Within a selected image, we iterate through input features; $p^1$ refers to the probability of masking a single pixel and $p^2$ refers to the probability of creating a non-informative square. For concreteness, $p^2$ is selected from a range $(0,1)$ for all mini-batches. Then, the perturbation is defined as follows. First, for each mini-batch, we draw $p^1$ from Uniform$(0, p^1_{\text{max}})$ distribution and set input pixels to $0$ with a probability of $p^1$. Second, with a probability of $p^2$ for each selected pixel, we create a non-informative square of $0$'s, with a randomly chosen side length in the interval $[1, s_{\text{max}}]$. This specific definition of perturbations is of heuristic nature and we found it to work well in our experiments. It is by no means unique or optimal and many other design choices are possible. See details in the Algorithm \autoref{alg:fpa}.

\begin{algorithm}
    \caption{Feature Perturbation Augmentation in a Selected Mini-batch}
    \begin{algorithmic}
    \Require $K$ samples $\mathbf{X}^k_{w, h, c}$ for $k = 0, \ldots, K$, where $\mathbf X^k$ is of dimension $W\times H\times C$.
    \Require s\ts{max} $< \min(W,H)$,\; $p^1_{\text{max}} \in (0,1)$, $p^2 \in (0,1)$

    \State Set $p^1 \sim \textnormal{Uniform}(0, p^1_{\text{max}})$
    \NoDo
        \For{$k \gets 0$ to $K$}
            \For{$w \gets 0$ to $W$}
                \For{$h \gets 0$ to $H$}
                    \State With $p^1$, $\mathbf{X}^k_{w, h} \gets 0$ (i.e., a non-informative value).
                    \State Set $s \gets \{1,2,\ldots,s_{\text{max}}\}$
                    %\State With $p^2$, create a square at $X_{i,j}$ of dimension $s\times s$ of non-informative values (e.g., 0).
                    \State With $p^2$, $\mathbf{X}^k_{w:(w+s), h:(h+s)} \gets 0$ (i.e., a $s\times s$ square of non-informative values). 
                \EndFor
            \EndFor
        \EndFor
    \end{algorithmic}
    \label{alg:fpa}
\end{algorithm}

Besides masking pixels with zeros, many different schemes of masking are possible, such as setting pixels to random, minimum, or maximum values or applying blurring, bokeh, or other filters. Similar to an ambiguous problem of choosing a non-informative baseline for integrated gradients \cite{sundararajan2017axiomatic, sturmfels2020visualizing}, perturbation should reflect the application domain. Beyond the specific implementation (Algorithm \autoref{alg:fpa}), our proposal is generally applicable. To make perturbation-based interpretability evaluate more reliable, one should include similarly perturbed samples during training.

% The underlying issue is that it is in general not clear which choice most effectively removes information. For instance, a grey box could effectively remove evidence for a red car but much less so for a grey one. This issue is closely related to the problem of choosing a non-informative baseline for integrated gradients \cite{sundararajan2017axiomatic,sturmfels2020visualizing}.

\subsection*{Datasets and Importance Estimators}

We demonstrate our approach using two popular deep learning architectures and three datasets, namely the ResNet-50 \citep{he2016deep} architecture trained on ImageNet \citep{deng2009imagenet} and Food101 \citep{bossard2014food}, and ResNet-18 trained on CIFAR-10 \citep{cifar}, see \ref{details} for details concerning the datasets and training procedure. We compare the following four importance estimators: vanilla gradient (VG) \citep{simonyan2014deep}, integrated gradients (IG) \citep{sundararajan2017axiomatic}, SmoothGrad (SG) \citep{smilkov2017smoothgrad} and squared SmoothGrad (SQ-SG) \citep{hooker2019benchmark}. 

These methods output three-dimensional maps of importance scores (height, width, and color channels). To obtain two-dimensional maps for the pixel-wise perturbation, we explore two variants: \textit{unsigned} and \textit{signed}. First, for the unsigned estimators, we sum the absolute values of color channels, which are denoted by the subscript \ts{abs}. In this case, very small values ($\gtrsim 0$) have minimal influence on that prediction. Second, for the signed estimators, we multiply the raw importance scores element-wise with the input image \cite{shrikumar2017learning}, (indicated by a prime) and sum over the resulting color channels (indicated by a subscript \ts{sum}). For example, VG$'$\ts{sum}. Negative importance scores from the signed estimators may imply \emph{counter-evidence} for the predicted class. IG includes a multiplication with the input by definition and will therefore not appear primed. 

We note that VG$'$ is equivalent to $\epsilon$-LRP \cite{bach2015pixel} when non-linearities are ReLUs \cite{ancona2017towards}, as in the ResNet architectures. Furthermore, multiplication with the input should be used with caution, since important pixels can be assigned, erroneously, a zero or low score when they are multiplied with a zero or low pixel value \cite{brocki2021input, sturmfels2020visualizing}. See \ref{estimators} for more details about the importance estimators.

\subsection*{Fidelity of Importance Estimators}

\begin{wrapfigure}{R}{0.4\textwidth}
\centering
\includegraphics[width=0.4\textwidth]{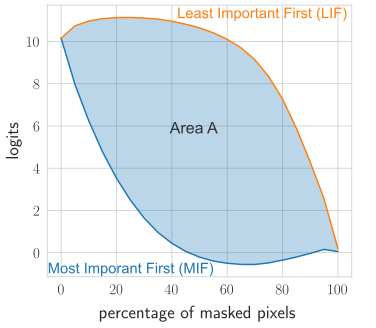}
\caption{The fidelity metric $A$ is defined as the area between the LIF (orange) and MIF (blue) curves. Importance estimators with larger $A$ are considered to explain the model more accurately.} 
\label{defineA}
\end{wrapfigure}

\begin{figure}[ht!]
\centering
\includegraphics[width=1.0\textwidth]{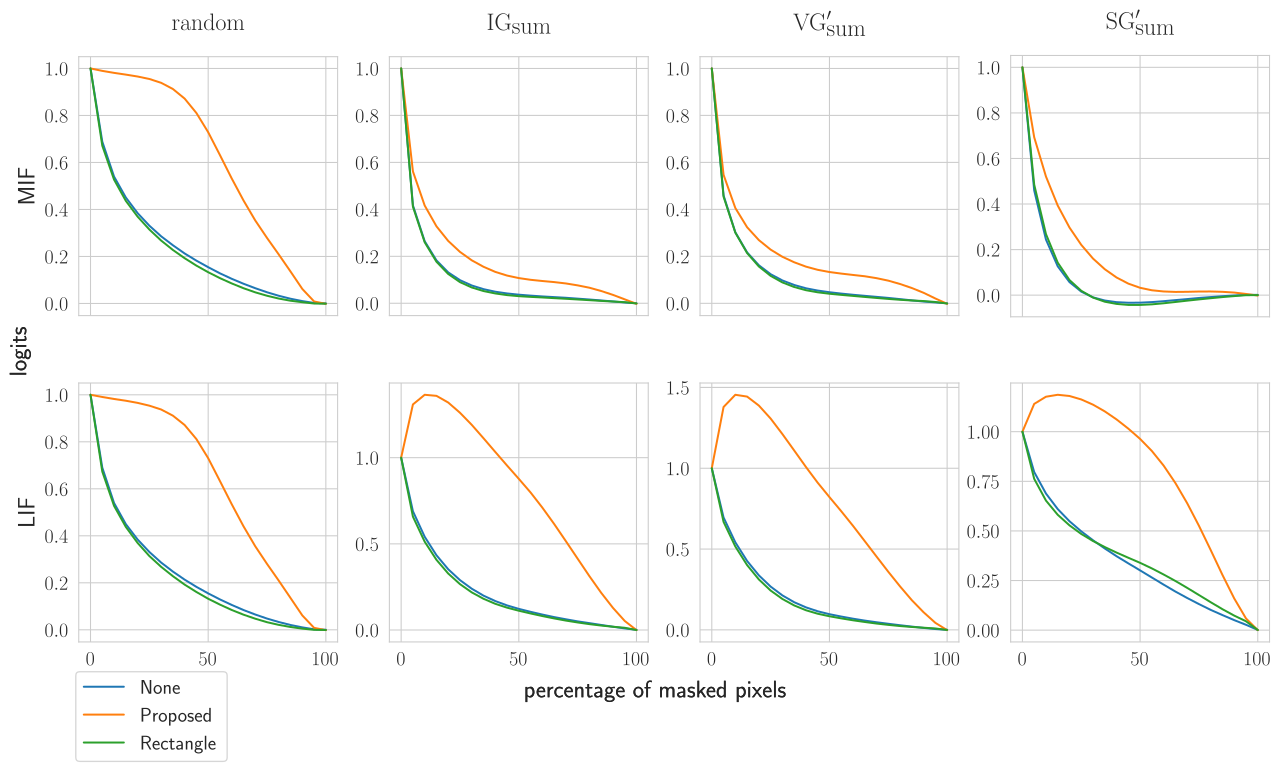}
\caption{Perturbation curves for the ResNet-50 trained on Food101 data with importance scores obtained using \textit{signed importance estimators}, i.e. both positive and negative importance scores available. ``Rectangle" refers to the data augmentation proposed in \cite{zhong2020random}. ``Random" (far left) means that importance scores are randomly assigned and the initial, unperturbed, logits have been normalized to one.} 
\label{food_signed}
\end{figure}

Perturbation-based evaluation methods are both intuitive and popular \cite{samek2016evaluating,petsiuk2018rise,kindermans2017learning}. For our evaluation of fidelity, we create \emph{perturbation curves} of changes in logits, i.e. pre-softmax activations in the prediction vector, with respect to an increasing amount of perturbation (e.g., \cref{defineA}). When needed for comparison, logits are normalized against the original model prediction without any masked pixel, e.g. in \cref{food_signed}. A given importance estimator computes a set of importance scores for input pixels, which indicate how much each pixel contributed to the final prediction. Then, input pixels are perturbed in order of either the most important first (MIF) or the least important first (LIF) \cite{brocki2021fidelity}. In the case of signed importance scores, the ranking goes from the highest positive values to the lowest negative ones (MIF), or reversely (LIF). Therefore, the lowest negative importance scores, which are ranked first in LIF, may indicate strong counter-evidence for the predicted class.

When the importance estimator deems a certain feature important, ideally the removal of this feature would strongly decrease the associated logit. A greater logit decrease would imply that a chosen feature is more important for the model's prediction. Inversely, if a feature has a low-ranked importance score, its removal would lead to a minimal accuracy decrease (unsigned estimators) or potentially an accuracy increase for removing counter-evidence (signed estimators). To obtain the final perturbation curves, we average normalized logits over all 10,000 samples in the test set for CIFAR-10. For ImageNet and Food101, we average over a randomly selected subset of 5,000 samples from the test set.
 
In order to combine these two aspects, we use the area $A$ between the MIF and LIF curves (\cref{defineA}) as a metric to measure the relative fidelity of importance estimators. A small area under the MIF curve indicates that the estimator is good at detecting features that are important evidence for a given class. A large area under the LIF curve, on the other hand, means that the estimator can reliably find unimportant features; negative importance scores (e.g., gradients) would imply counter-evidence for the prediction. With $A$ as fidelity metric, we, therefore, consider importance estimators with large $A$ to be overall superior to ones with lower $A$.

%The use of the area $A$ between the MIF and LIF curves as the fidelity of importance estimators mitigates a potential caveat of our method; namely, that some importance estimators and their pattern of feature perturbation may benefit more from data augmentation than others. We have observed that random erasing \cite{zhong2020random} sometimes has very limited influence on the perturbation curves (e.g., \cref{food_signed}). We hypothesize that this may happen since the spatial distribution of masked pixels in that augmentation is widely different from the one the model encounters when the images are perturbed according to MIF/LIF orders. A similar phenomenon could occur with the proposed FPA if the spatial distribution of masked pixels for one estimator is more similar to the augmentation than for another one. By measuring the fidelity with $A$ this effect should, at least partially, cancel out since one would expect both the areas under the LIF and MIF curves to increase due to it. 

\section{Results}\label{results}
\subsection{Accuracy of models}

We applied the proposed FPA on three datasets; the ImageNet \cite{deng2009imagenet}, Food101 \cite{bossard2014food}, and CIFAR-10 \cite{cifar}. The ResNet-50 \cite{he2016deep} architecture is used for ImageNet and Food101 and ResNet-18 for CIFAR-10. Desirable parameters for FPA significantly improve the model's robustness while maintaining its accuracy. In order to find a good set of parameters we performed a partial grid search, keeping $p^2$ and $s_{\text{max}}$ fixed and varying $p$ in the range $[0.2,0.5]$ and $p^1_{\text{max}}$ in $[0.1,0.4]$ using $0.1$ and $0.5$ steps for Food101 and CIFAR-10, respectively. Due to restrictions in computing resources, we did not include $p^2$ and $s_{\text{max}}$ in the grid search. FPA parameters for the ImageNet were set to the same parameters selected for Food101. For the augmentation of CIFAR-10, we chose $p^1_{\text{max}}=0.25,\;p^2=0.1,\; s_{\text{max}}=3$ and for ImageNet and Food101  $p^1_{\text{max}}=0.3,\;p^2=0.01,\;s_{\text{max}}=10$. We set $p=0.5$ for all three datasets. The chosen parameter values present a good trade-off between the increased robustness and performance of the model. 

Evaluated on the same images that are used to obtain the perturbation curves, the models trained on CIFAR-10 have an accuracy of $93.0\%$, $92.7\%$ and $93.1\%$ for no augmentation, proposed FPA and ``random erasing" \cite{zhong2020random} (denoted as ``Rectangle'' in our figures), respectively. In the same order, the models trained on ImageNet have an accuracy of $76.2\%$, $74.7\%$ and $75.9\%$ and on Food101 $83.5\%, 81.5\%$ and $83.5\%$.% In contrast to other augmentation methods, the aim of FPA is not to increase performance but instead to improve interpretability by making the evaluation of importance estimators more reliable. In that sense, we are trading some of the model performance for improved interpretability.

\begin{figure}[ht!]
\centering
\includegraphics[width=0.4\textwidth]{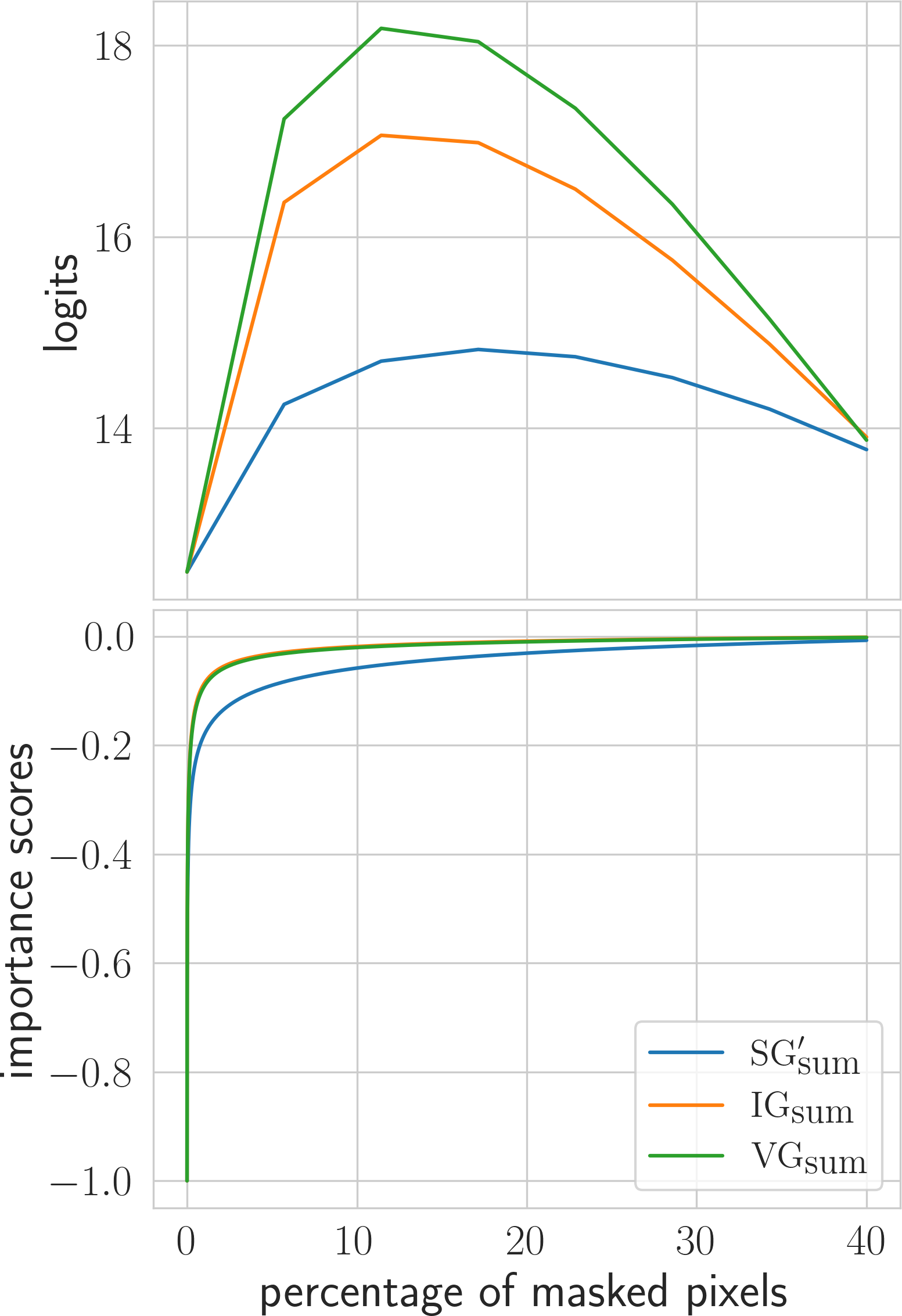}
\caption{Comparison of the model output and importance scores when FPA is used. Masking pixels with negative importance scores coincides with increasing logit values. \emph{Top:} LIF perturbation curves of logits for Food101. \emph{Bottom:} Importance scores of pixels that are masked in the LIF order.} 
\label{food_compare}
\end{figure}
%The curve has been obtained by flattening the heat map and plotting the importance scores in LIF order, with the x-axis indicating their position in the ranking. According to the x-axis, compare the importance scores of a group of pixels and the influence its masking had on the model output.

\subsection{Perturbation curves}

Once the model is trained with or without data augmentation, vanilla gradient (VG), integrated gradient (IG), Smoothgrad (SG), and squared SmoothGrad (SQ-SG) are applied to obtain matrices of importance scores. We also calculate perturbation curves as described in \cref{methods}. In \cref{food_signed}, the MIF perturbation curves (top row) fall off slower when the model was trained with the proposed FPA, compared to those with no augmentation or Rectangle augmentation \cite{zhong2020random}. For LIF perturbation curves (bottom row, \cref{food_signed}) the logits initially increase before decreasing if FPA is used. In contrast, without any augmentation or with ``Rectangle'' augmentation, the logits immediately and rapidly decrease. The random baseline (where importance scores are randomly assigned to pixels) does not exhibit the early increase of logits.

These operating characteristics are expected when the influence of artifacts is removed, or at least strongly reduced, by our proposed augmentation. Generally, some of the logit decrease is expected to be due to perturbation artifacts, thus removing perturbation artifacts would delay the logit decrease. In the LIF curves in \cref{food_signed}, pixels with large negative importance scores are masked first which removes \textit{counter-evidence}. This is highlighted in \cref{food_compare}, which demonstrates that masking pixels with a negative importance score coincides with an increase in the logit values (see \cref{cifar_imgnet_compare} for equivalent graphs for CIFAR-10 and ImageNet). In contrast, without FPA, this effect is suppressed and we hypothesize that this is due to perturbation artifacts disturbing the model. This would lead to a net decrease of the logits, despite pixels with negative scores, i.e. counter-evidence, being masked. This behavior is consistent across the three considered datasets and architectures (see \cref{summary_signed,summary_unsigned}).

%We also observe that the LIF curves in \cref{food_compare} decrease even when masking pixels with importance scores that are slightly negative, whereas one may have expected for the model output too slowly increase in such a case. One possible explanation is that when the number of masked pixels increases, also in the case of training with FPA perturbation artifacts are present which cause the unexpected decrease. Another explanation is that for a larger fraction of masked pixels interactions between them occur that the importance estimator did not account for so that the  scores are not reliable anymore. 

\subsection{Comparing fidelity of importance estimators}

We compare the fidelity $A$ of importance estimators (\cref{table:3,table:1,table:2}) considering two different settings. In the first setting, we take into account the magnitude of importance scores and disregard whether their signs correctly indicate evidence or counter-evidence for the prediction. Non-negative importance scores are obtained from the \emph{unsigned estimators}: IG\ts{abs}, VG\ts{abs}, VG$'$\ts{abs}, SG\ts{abs}, SG$'$\ts{abs} and SQ-SG\ts{sum}. Among those that rank pixels purely by the magnitude of importance scores, we find that across all three considered datasets, and regardless of the augmentation, SQ-SG\ts{sum} consistently performs best and VG\ts{abs} worst.

\begin{table}[ht]
\centering
\caption{The fidelity of importance estimators $A$ (the area between LIF and MIF perturbation curves), measured on the ResNet-50 trained on Food101 with $95\%$ confidence intervals.
}
\label{table:3}
\begin{tabular}{l c c c c c c} 

 Aug. & Random & IG\ts{sum} & IG\ts{abs}  & VG\ts{abs} & VG$'$\ts{sum}\\\midrule

 None & $0.0\pm 0.8$ & $11.6\pm 0.7$ & $22.6\pm 0.8$ & $15.5\pm 0.8$ & $8.7\pm 0.6$ \\ 
 Proposed & $0.0\pm 0.6$ & $67.7\pm 1.0$ & $29.4\pm 0.6$ & $24.8\pm 0.7$ & $66.0\pm 1.0$ \\ \\

 Aug. & VG$'$\ts{abs}  &SG\ts{abs} &  SG$'$\ts{sum} &SG$'$\ts{abs}& SQ-SG\ts{sum}\\\midrule

 None & $15.5\pm 0.8$ & $22.2\pm 0.8$ & $30.3\pm 0.8$ & $23.7\pm 0.7$& $\mathbf{31.2\pm 0.8}$\\ 
 Proposed & $24.8\pm 0.7$ & $28.9\pm 0.7$ & $\mathbf{69.4\pm 0.9}$ & $27.0\pm 0.6$ & $31.3\pm0.7$\\ 
\end{tabular}
\end{table}

For the second setting, we consider signed importance scores from the \emph{signed estimators}; namely, IG\ts{sum}, VG$'$\ts{sum} and SG$'$\ts{sum}, which multiply the gradients and input element-wise\footnote{Notice that the gradients by themselves, without multiplying with the input, can not correctly infer the sign of importance scores. This becomes clear when one considers a linear model $f(x) = \omega x$, the gradient $\frac{\partial f}{\partial x}=\omega$ does not contain information on the sign of $f$ if $x$ can be negative. Multiplying with the input yields $x \frac{\partial f}{\partial x}$ and thus VG$'$ gives the correct sign in the linear case. IG has a multiplication with the input built into its definition.}. With FPA, the signed estimators consistently outperform the unsigned ones across all three datasets, with IG\ts{sum} and SG$'$\ts{sum} in the leading positions. Intuitively, this makes sense since the signed estimators contain more information than the unsigned ones, allowing them to describe the models' predictions more accurately.
% Without FPA, the unsigned SQ-SG\ts{sum} outperforms other methods on Food101 and ImageNet, and SG$'$\ts{sum} outperforms others on CIFAR-10. 
%This indicates that using FPA makes the obtained ranking of importance estimators more reliable.

\begin{wrapfigure}{R}{0.4\textwidth}
\centering
\includegraphics[width=0.4\textwidth]{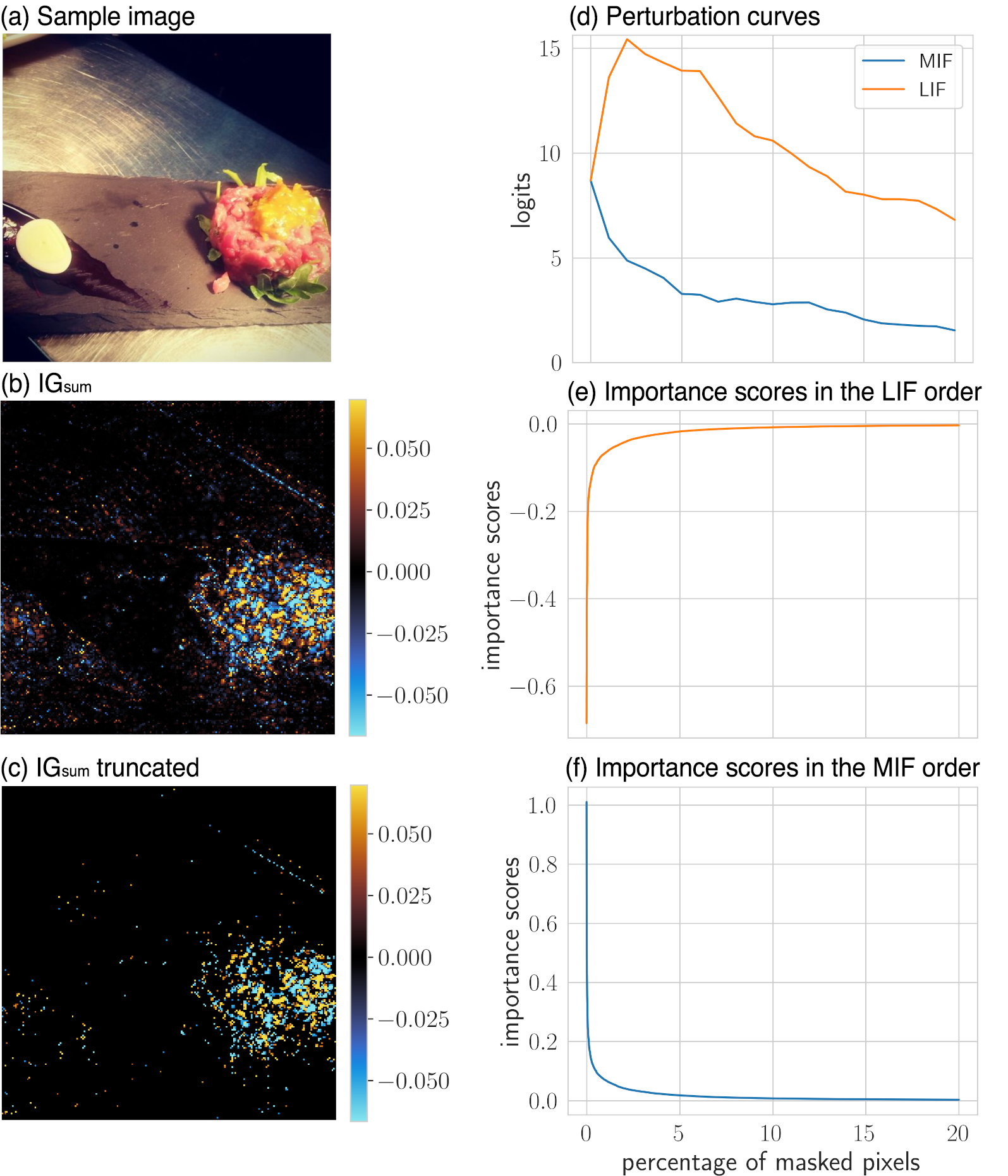}
\caption{Explainability of the sign of importance scores and perturbation curves, using a sample image from Food101 (correctly classified as ``beef tartare''). In (c), pixels with IG\ts{sum} scores lower than the 98th percentile are set to zero, and the importance scores are clipped at the 99th percentile.} 
\label{food_fluctuation}
\end{wrapfigure}
% and heat maps of importance scores obtained with IG\ts{sum}, where the upper heat map is the original one and in the lower one all pixels with scores lower than the 98th percentile are set to zero. To increase contrast of the heat maps the importance scores are clipped at the 99th percentile. Right: The top graph shows the corresponding MIF and LIF perturbation curves with FPA used during training and the lower ones have been obtained in the same manner as the curves on the bottom of \cref{food_compare}, here for LIF and MIF.
\subsection{Fluctuations in the sign of importance scores}
The high-frequency fluctuations in the sign of importance scores have previously been attributed to the inability of importance estimators to predict the sign accurately  \cite{ancona2017towards,samek2021explaining}. However, our results demonstrate that the observed fluctuations describe the model characteristics rather accurately if perturbation artifacts are suppressed by FPA. For illustrative purposes, we focus on a randomly chosen sample in \cref{food_fluctuation}. The trained ResNet-50 model is generally focusing on the object of interest (correctly predicted as ``beef tartare''), but we also observe that within that object, the sign of importance scores fluctuates. The distribution of negative (\cref{food_fluctuation}(e)) and positive (\cref{food_fluctuation}(f)) importance scores are highly concentrated ($\sim2\%$ of all pixels each).

We further created a truncated heatmap (\cref{food_fluctuation}(c)) where only the highest and lowest importance scores are plotted (thresholding at the 98th percentile), highlighting only those pixels whose removal causes the initial increase and decrease in logits in \cref{food_fluctuation}(d). The vast majority of those pixels are within the region of interest (ROI). Finally, the right-side plots (\cref{food_fluctuation}(d-f)) show that the initial increase (or decrease) of logits in the LIF (or MIF) curve coincides with masking large negative (or positive) values. Taken together, these observations imply that the sign fluctuations may actually be meaningful.

\section{Discussion}\label{discussion}
% this paragraph should be merged with Conclusion, currently its largely redundant 
% We introduce an intuitive technique to augment the data that can improve evaluation of post-hoc interpretability methods which estimate importance scores of input pixels. Perturbation-based evaluation of such importance estimators is a mainstay, despite masking pixels are known to cause the model to erratically behave due to the out-of-distribution problem. In the proposed feature perturbation augmentation (FPA), randomly selected samples undergo the similar perturbation process where some pixels are masked to non-informative values. When applied to three datasets and four post-hoc explanation methods, training DNNs with FPA help protect against perturbation artifacts and make their evaluation more reliable.

As this work has been focused on the interpretability and trustworthiness of deep learning models, there still is room for performance improvement. Training the models with FPA leads to a slight decrease in performance compared to training them without augmentation. This might be due to ``augment ambiguity" \cite{wei2020circumventing}, which occurs when the annotated class is not recognizable anymore as a result of an augmentation. 
%Generally, there exists a trade-off between interpretability and accuracy \citep{dziugaite2020tradeoff}. 
On the other hand, random erasing, which is closely related to FPA, reported an increase in the model performance in certain settings \citep{zhong2020random}. In future work, we plan to explore variations of FPA and training schemes to mitigate the performance loss, but also point out that trading some accuracy for increased interpretability can be worthwhile.

Our results show that VG, SG, and IG outperform a random baseline across all three considered datasets. Comparing our ranking of importance estimators to those in the literature we find that, in contrast to our results, ROAR \cite{hooker2019benchmark} reports that VG, SG, and IG perform worse than a random baseline. Several other evaluation methods, however, report that IG performs better than a random baseline as well \cite{kapishnikov2019xrai,rieger2020irof,ancona2017towards}. We hypothesize that the counter-intuitive results of ROAR arise because, by measuring the accuracy of the retrained models (cf. \cref{related}), one gauges how much information has been removed and was not available for retraining the model. In general, this is different from gauging how much information has been removed \textit{that the original model relies on}. Our proposed method avoids this pitfall by training the model only once with FPA and then using the same model for all following evaluations. Similar concerns with ROAR have been raised in \cite{nguyen2020quantitative}.
% \cite{samek2021explaining} claims that IG performs well because it constructs adversarial examples, which cause the rapid performance decrease. Our results seem to suggest otherwise since evidence and counter-evidence are both correctly identified by IG (e.g., \cref{food_fluctuation}).
 
Furthermore, the performed experiments bring out an interesting insight into the frequently observed fluctuation of the sign of importance scores obtained from gradient-based or LRP methods. Often, these fluctuations have been attributed to a failure of post-hoc explanations to correctly predict the sign and are considered an issue that requires fixing  \cite{kapishnikov2019xrai,ancona2017towards,samek2021explaining}. Our results imply that these fluctuations may not be a bug but a feature. To arrive at its predictions, the model often focuses on the ROI, but within, it may find rapidly alternating evidence and counter-evidence on the pixel-level. Deep neural networks and humans appear to arrive at their predictions in very different ways and we would therefore caution against forcing post-hoc explanations to align with human expectations.

Concerning possible extensions of our work, notice that since FPA randomly selects pixels to construct augmented training samples it can be expected to only increase a model's robustness against such random perturbations. To increase the robustness against adversarial perturbations our work could be extended by additionally performing adversarial training \cite{goodfellow2014explaining, madrytowards}, which would allow to rule out perturbation artifacts during evaluation with even higher confidence. 

% this is mentioned in method section:
%Lastly, we note that although FPA has been introduced here for the case of perturbing pixels with constant values, it can be applied to other perturbation methods. For instance, FPA can be used to make Performance Information Curves \cite{kapishnikov2019xrai} more reliable by augmenting the training with images in which a bokeh filter is partially applied. The only requirement is that the perturbation performed to augment images needs to reflect the perturbation applied during evaluation. 
%Instead, to obtain such an alignment of the human and machine process of reasoning, it would be required to design inherently interpretable models \cite{rudin2019stop} with built-in alignment.

\section{Conclusions}\label{conclusions}
We demonstrate that the perturbation-based evaluation of importance estimators for deep neural networks (DNNs) can be made more reliable by our proposed feature perturbation augmentation (FPA). Our experiments using the Most Important First (MIF) and Least Important First (LIF) perturbation curves demonstrate that our FPA training makes the DNNs more robust against perturbations and this effectively mitigates the issue of confounding artifacts introduced by masking input pixels during interpretability evaluation. To the best of our knowledge, FPA is the first data augmentation technique specifically developed to improve the evaluation of interpretability methods in DNNs.

%This claim is substantiated by our observation that when presumed counter-evidence is masked, the expected effect of an \textit{increase} in logit values only occurs when we used FPA during training. Without FPA, for all considered importance estimators the logits always decrease, which can be naturally explained by the disturbance that the masking of pixels causes in the model. Interestingly, our results suggest that the sign fluctuations frequently observed in heatmaps of importance scores are not due to inaccuracies of the explanation methods but, more likely, they faithfully describe the model's operating characteristics. 

\section*{Acknowledgements}
This work was funded by the ERA-Net CHIST-ERA grant [CHIST-ERA-19-XAI-007] long term challenges in ICT project INFORM (ID: 93603), by the National Science Centre (NCN) of Poland [2020/02/Y/ST6/00071]. This research was carried out with the support of the Interdisciplinary Centre for Mathematical and Computational Modelling University of Warsaw (ICM UW) under computational allocation no GDM-3540; the NVIDIA Corporation’s GPU grant; and the Google Cloud Research Innovators program.

%% If you have bibdatabase file and want bibtex to generate the
%% bibitems, please use
%%
 \bibliographystyle{elsarticle-num} 
 \bibliography{sample}

\begin{thebibliography}{10}
\expandafter\ifx\csname url\endcsname\relax
  \def\url#1{\texttt{#1}}\fi
\expandafter\ifx\csname urlprefix\endcsname\relax\def\urlprefix{URL }\fi
\expandafter\ifx\csname href\endcsname\relax
  \def\href#1#2{#2} \def\path#1{#1}\fi

\bibitem{samek2019explainable}
W.~Samek, G.~Montavon, A.~Vedaldi, L.~K. Hansen, K.-R. M{\"u}ller, Explainable
  AI: interpreting, explaining and visualizing deep learning, Vol. 11700,
  Springer Nature, 2019.

\bibitem{simonyan2014deep}
K.~Simonyan, A.~Vedaldi, A.~Zisserman, Deep inside convolutional networks:
  Visualising image classification models and saliency maps, in: In Workshop at
  International Conference on Learning Representations, Citeseer, 2014.

\bibitem{smilkov2017smoothgrad}
D.~Smilkov, N.~Thorat, B.~Kim, F.~Vi{\'e}gas, M.~Wattenberg, Smoothgrad:
  removing noise by adding noise, arXiv preprint arXiv:1706.03825 (2017).

\bibitem{sundararajan2017axiomatic}
M.~Sundararajan, A.~Taly, Q.~Yan, Axiomatic attribution for deep networks, in:
  International conference on machine learning, PMLR, 2017, pp. 3319--3328.

\bibitem{samek2016evaluating}
W.~Samek, A.~Binder, G.~Montavon, S.~Lapuschkin, K.-R. M{\"u}ller, Evaluating
  the visualization of what a deep neural network has learned, IEEE
  transactions on neural networks and learning systems 28~(11) (2016)
  2660--2673.

\bibitem{petsiuk2018rise}
V.~Petsiuk, A.~Das, K.~Saenko, Rise: Randomized input sampling for explanation
  of black-box models, arXiv preprint arXiv:1806.07421 (2018).

\bibitem{kindermans2017learning}
P.-J. Kindermans, K.~T. Sch{\"u}tt, M.~Alber, K.-R. M{\"u}ller, D.~Erhan,
  B.~Kim, S.~D{\"a}hne, Learning how to explain neural networks: Patternnet and
  patternattribution, arXiv preprint arXiv:1705.05598 (2017).

\bibitem{hooker2019benchmark}
S.~Hooker, D.~Erhan, P.-J. Kindermans, B.~Kim, A benchmark for interpretability
  methods in deep neural networks, Advances in neural information processing
  systems 32 (2019).

\bibitem{fong2017interpretable}
R.~C. Fong, A.~Vedaldi, Interpretable explanations of black boxes by meaningful
  perturbation, in: Proceedings of the IEEE international conference on
  computer vision, 2017, pp. 3429--3437.

\bibitem{goodfellow2014explaining}
I.~J. Goodfellow, J.~Shlens, C.~Szegedy, Explaining and harnessing adversarial
  examples, arXiv preprint arXiv:1412.6572 (2014).

\bibitem{cifar}
A.~Krizhevsky, Learning multiple layers of features from tiny images,
  University of Toronto Technical Report (2009).

\bibitem{bossard2014food}
L.~Bossard, M.~Guillaumin, L.~V. Gool, Food-101--mining discriminative
  components with random forests, in: European conference on computer vision,
  Springer, 2014, pp. 446--461.

\bibitem{deng2009imagenet}
J.~Deng, W.~Dong, R.~Socher, L.-J. Li, K.~Li, L.~Fei-Fei, Imagenet: A
  large-scale hierarchical image database, in: 2009 IEEE conference on computer
  vision and pattern recognition, Ieee, 2009, pp. 248--255.

\bibitem{samek2021explaining}
W.~Samek, G.~Montavon, S.~Lapuschkin, C.~J. Anders, K.-R. M{\"u}ller,
  Explaining deep neural networks and beyond: A review of methods and
  applications, Proceedings of the IEEE 109~(3) (2021) 247--278.

\bibitem{rudin2022interpretable}
C.~Rudin, C.~Chen, Z.~Chen, H.~Huang, L.~Semenova, C.~Zhong, Interpretable
  machine learning: Fundamental principles and 10 grand challenges, Statistics
  Surveys 16 (2022) 1--85.

\bibitem{saporta2022benchmarking}
A.~Saporta, X.~Gui, A.~Agrawal, A.~Pareek, S.~Q. Truong, C.~D. Nguyen, V.-D.
  Ngo, J.~Seekins, F.~G. Blankenberg, A.~Y. Ng, et~al., Benchmarking saliency
  methods for chest x-ray interpretation, Nature Machine Intelligence (2022)
  1--12.

\bibitem{brocki2022evaluation}
L.~Brocki, W.~Marchadour, J.~Maison, B.~Badic, P.~Papadimitroulas, M.~Hatt,
  F.~Vermet, N.~C. Chung, Evaluation of importance estimators in deep learning
  classifiers for computed tomography, in: International Workshop on
  Explainable, Transparent Autonomous Agents and Multi-Agent Systems, Springer,
  2022, pp. 3--18.

\bibitem{zhou2016learning}
B.~Zhou, A.~Khosla, A.~Lapedriza, A.~Oliva, A.~Torralba, Learning deep features
  for discriminative localization, in: Proceedings of the IEEE conference on
  computer vision and pattern recognition, 2016, pp. 2921--2929.

\bibitem{bach2015pixel}
S.~Bach, A.~Binder, G.~Montavon, F.~Klauschen, K.-R. M{\"u}ller, W.~Samek, On
  pixel-wise explanations for non-linear classifier decisions by layer-wise
  relevance propagation, PloS one 10~(7) (2015) e0130140.

\bibitem{bhatt2020evaluating}
U.~Bhatt, A.~Weller, J.~M. Moura, Evaluating and aggregating feature-based
  model explanations, arXiv preprint arXiv:2005.00631 (2020).

\bibitem{ancona2017towards}
M.~Ancona, E.~Ceolini, C.~{\"O}ztireli, M.~Gross, Towards better understanding
  of gradient-based attribution methods for deep neural networks, arXiv
  preprint arXiv:1711.06104 (2017).

\bibitem{dabkowski2017real}
P.~Dabkowski, Y.~Gal, Real time image saliency for black box classifiers,
  Advances in neural information processing systems 30 (2017).

\bibitem{kapishnikov2019xrai}
A.~Kapishnikov, T.~Bolukbasi, F.~Vi{\'e}gas, M.~Terry, Xrai: Better
  attributions through regions, in: Proceedings of the IEEE/CVF International
  Conference on Computer Vision, 2019, pp. 4948--4957.

\bibitem{rong2022consistent}
Y.~Rong, T.~Leemann, V.~Borisov, G.~Kasneci, E.~Kasneci, A consistent and
  efficient evaluation strategy for attribution methods, in: International
  Conference on Machine Learning, PMLR, 2022, pp. 18770--18795.

\bibitem{rieger2020irof}
L.~Rieger, L.~K. Hansen, Irof: a low resource evaluation metric for explanation
  methods, arXiv preprint arXiv:2003.08747 (2020).

\bibitem{yeh2019fidelity}
C.-K. Yeh, C.-Y. Hsieh, A.~Suggala, D.~I. Inouye, P.~K. Ravikumar, On the (in)
  fidelity and sensitivity of explanations, Advances in Neural Information
  Processing Systems 32 (2019).

\bibitem{dasgupta2022framework}
S.~Dasgupta, N.~Frost, M.~Moshkovitz, Framework for evaluating faithfulness of
  local explanations, arXiv preprint arXiv:2202.00734 (2022).

\bibitem{shorten2019survey}
C.~Shorten, T.~M. Khoshgoftaar, A survey on image data augmentation for deep
  learning, Journal of big data 6~(1) (2019) 1--48.

\bibitem{zhong2020random}
Z.~Zhong, L.~Zheng, G.~Kang, S.~Li, Y.~Yang, Random erasing data augmentation,
  in: Proceedings of the AAAI conference on artificial intelligence, Vol.~34,
  2020, pp. 13001--13008.

\bibitem{dziugaite2020tradeoff}
G.~K. Dziugaite, S.~Ben-David, D.~M. Roy, Enforcing interpretability and its
  statistical impacts: Trade-offs between accuracy and interpretability, arXiv
  preprint arXiv:2010.13764 (2020).

\bibitem{sturmfels2020visualizing}
P.~Sturmfels, S.~Lundberg, S.-I. Lee, Visualizing the impact of feature
  attribution baselines, Distill 5~(1) (2020) e22.

\bibitem{he2016deep}
K.~He, X.~Zhang, S.~Ren, J.~Sun, Deep residual learning for image recognition,
  in: Proceedings of the IEEE conference on computer vision and pattern
  recognition, 2016, pp. 770--778.

\bibitem{shrikumar2017learning}
A.~Shrikumar, P.~Greenside, A.~Kundaje, Learning important features through
  propagating activation differences, in: International conference on machine
  learning, PMLR, 2017, pp. 3145--3153.

\bibitem{brocki2021input}
L.~Brocki, N.~C. Chung, Input bias in rectified gradients and modified saliency
  maps, in: 2021 IEEE International Conference on Big Data and Smart Computing
  (BigComp), IEEE, 2021, pp. 148--151.

\bibitem{brocki2021fidelity}
L.~Brocki, N.~C. Chung, Evaluation of interpretability methods and perturbation
  artifacts in deep neural networks (2022).

\bibitem{wei2020circumventing}
L.~Wei, A.~Xiao, L.~Xie, X.~Zhang, X.~Chen, Q.~Tian, Circumventing outliers of
  autoaugment with knowledge distillation, in: Computer Vision--ECCV 2020: 16th
  European Conference, Glasgow, UK, August 23--28, 2020, Proceedings, Part III,
  Springer, 2020, pp. 608--625.

\bibitem{nguyen2020quantitative}
A.-P. Nguyen, M.~R. Mart{\'\i}nez, On quantitative aspects of model
  interpretability, arXiv preprint arXiv:2007.07584 (2020).

\bibitem{madrytowards}
A.~Madry, A.~Makelov, L.~Schmidt, D.~Tsipras, A.~Vladu, Towards deep learning
  models resistant to adversarial attacks, in: International Conference on
  Learning Representations.

\end{thebibliography}

%% The Appendices part is started with the command \appendix;
%% appendix sections are then done as normal sections
\appendix

\newpage

%\label{appendix}
\renewcommand\thefigure{\thesection.\arabic{figure}}    
\setcounter{figure}{0}    

\renewcommand\thetable{\thesection.\arabic{table}}    
\setcounter{table}{0}    

\section{Datasets and Training Procedure}\label{details}

We demonstrate our approach using two popular deep learning architectures and three datasets, namely the ResNet-50 \cite{he2016deep} architecture trained on ImageNet \cite{deng2009imagenet} and Food101 \cite{bossard2014food}, and ResNet-18 trained on CIFAR-10 \cite{cifar}. We adapted the ResNet-18 architecture to be suitable for the smaller input dimensions of CIFAR-10 by setting the kernel size of the first convolutional layer to $(3,3)$ and stride and padding to $(1,1)$. The ResNet-50 model was trained on ImageNet for 90 epochs using the SGD optimizer with momentum $0.9$, weight decay $10^{-4}$ and  initial learning rate 0.1, which we reduced by a factor of 10 on epochs 30 and 60. On Food101, Resnet-50 was trained for 68 epochs with momentum set to 0.9, weight decay $5\times10^{-4}$ and an initial learning rate of 0.1, where a cosine annealing schedule was used to continuously reduce the learning rate to zero. The ResNet-18 model was trained for 40 epochs using the SGD optimizer with momentum $0.9$, weight decay $5\times 10^{-4}$, and initial learning rate 0.01, which we reduced by a factor of 10 on epoch 30. For CIFAR-10 and ImageNet, we scaled input images to the range $[-1,1]$ and for Food101 we performed a z-score normalization with mean and standard deviation from the training set. In all three cases, we performed a horizontal flip with a probability of $0.5$ to augment the data. 

\section{Importance Estimators}\label{estimators}

After the aforementioned models are trained on ImageNet \cite{deng2009imagenet}, Food101 \cite{bossard2014food}, and CIFAR-10 \cite{cifar} with and without data augmentation, we applied post-hoc interpretability methods that assign importance scores to input features. In particular, we compare the following four importance estimators 
\begin{itemize}
  \item \textbf{Vanilla gradient (VG)} \cite{simonyan2014deep}: Gradients of the class score $S_c$ with respect to input pixels $x_i$
  \begin{align*}
    \mathbf e = \frac{\partial S_c}{\partial x}
  \end{align*}
   The class score $S_c$ is the activation of the neuron in the prediction vector that corresponds to the predicted class $c$ by the original model. Gradients are calculated with respect to logits, i.e., pre-softmax activations.
  \item \textbf{Integrated gradient (IG)} \cite{sundararajan2017axiomatic}: Average over gradients obtained from inputs interpolated between a reference point $x^0$ and input $x$
  \begin{equation*}
\mathbf{e}=\left({x}-{x}^{0}\right) \times \sum_{k=1}^{m} \frac{\partial S_c\left({x}^{0}+\frac{k}{m}\left({x}-{x}^{0}\right)\right)}{\partial x} \times \frac{1}{m},
\end{equation*}
    where $x^0$ is chosen to be a black image and $m=200$.
  \item \textbf{Smoothgrad (SG)} \cite{smilkov2017smoothgrad}: Average over gradients obtained from inputs with injected noise
    \begin{equation*}
    \mathbf e =\frac{1}{n} \sum_{1}^{n} \hat{\mathbf e}\left(x+\mathcal{N}\left(0, \sigma^{2}\right)\right),
    \end{equation*}
    where $\mathcal{N}\left(0, \sigma^{2}\right)$ represents Gaussian noise with standard deviation $\sigma$, $\hat{\mathbf e}$ is obtained using vanilla gradient and $n=15$.
    \item \textbf{Squared SmoothGrad (SQ-SG)} \cite{hooker2019benchmark}: Variant of SmoothGrad that squares $\hat{\mathbf e}$ before averaging
    \begin{equation*}
    \mathbf e =\frac{1}{n} \sum_{1}^{n} \hat{\mathbf e}\left(x+\mathcal{N}\left(0, \sigma^{2}\right)\right)^2.
    \end{equation*}
\end{itemize}

Generally, importance scores are three-dimensional including height, width, and color channels. To reduce them to two-dimensional maps for feature perturbation, we compute the sum of the absolute values of color channels (\emph{unsigned}) and the sum of element-wise multiplication between the raw importance scores and the input image (\emph{signed}). See \cref{methods} for further details. 

\clearpage
\section{Supplementary Figures}
\begin{figure}[ht!]
\centering
\includegraphics[width=1.0\textwidth]{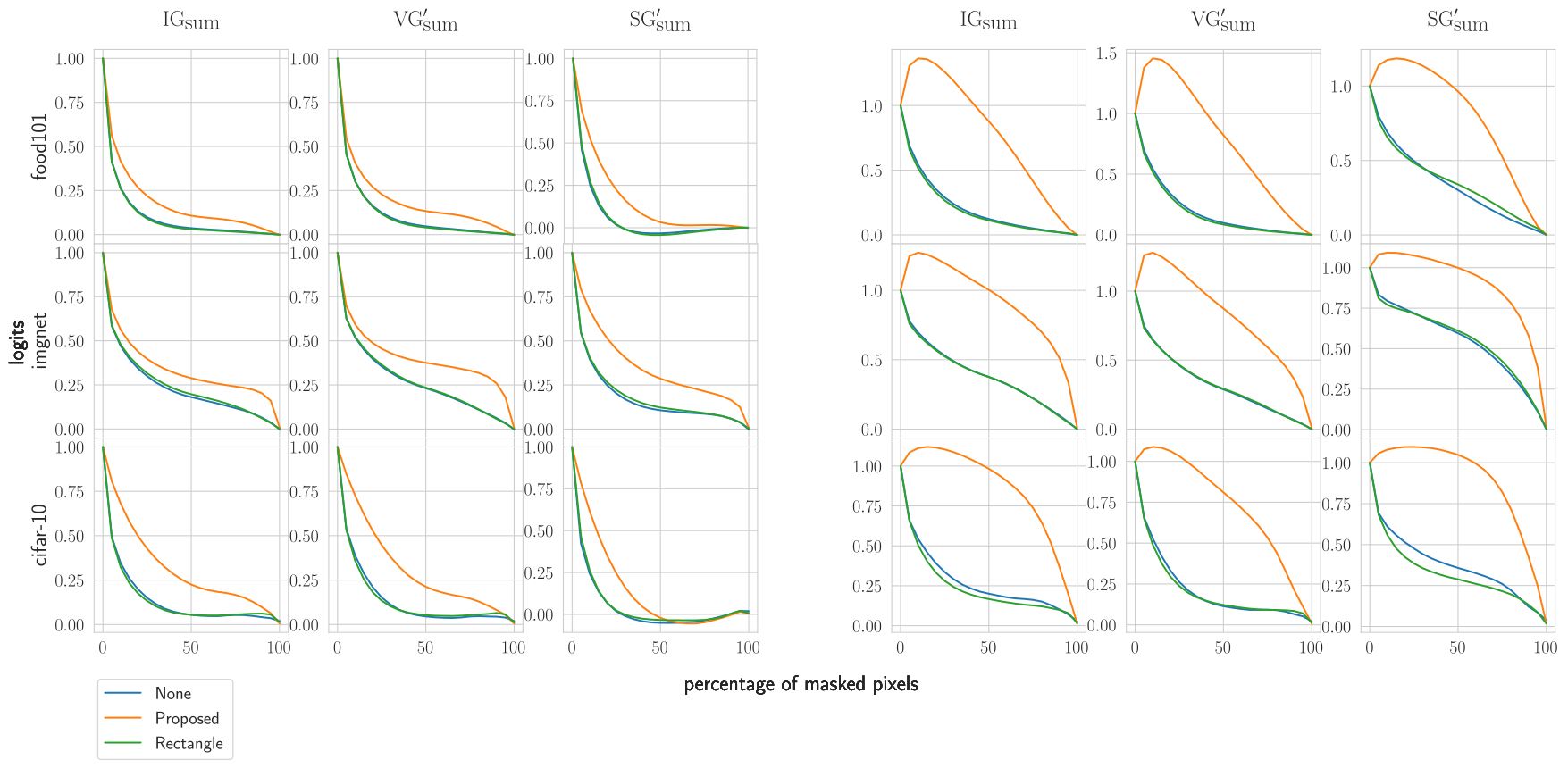}
\caption{Overview of perturbation curves for signed importance estimators for all three considered datasets. The raw importance scores are multiplied element-wise with the input image \cite{shrikumar2017learning}, indicated by a prime, and then followed by summing over the resulting color channels. By definition, IG includes a multiplication with the input already. The change in normalized logits is measured as a percentage of pixels that are masked according to the Most Important First (MIF; \emph{Left}) and the Least Important First (LIF; \emph{Right}). Note that masking pixels with negative importance scores can increase logits, as counter-evidence for the predicted class is removed from the input image.} 
\label{summary_signed}
\end{figure}

\begin{figure}[ht!]
\centering
\includegraphics[width=1.0\textwidth]{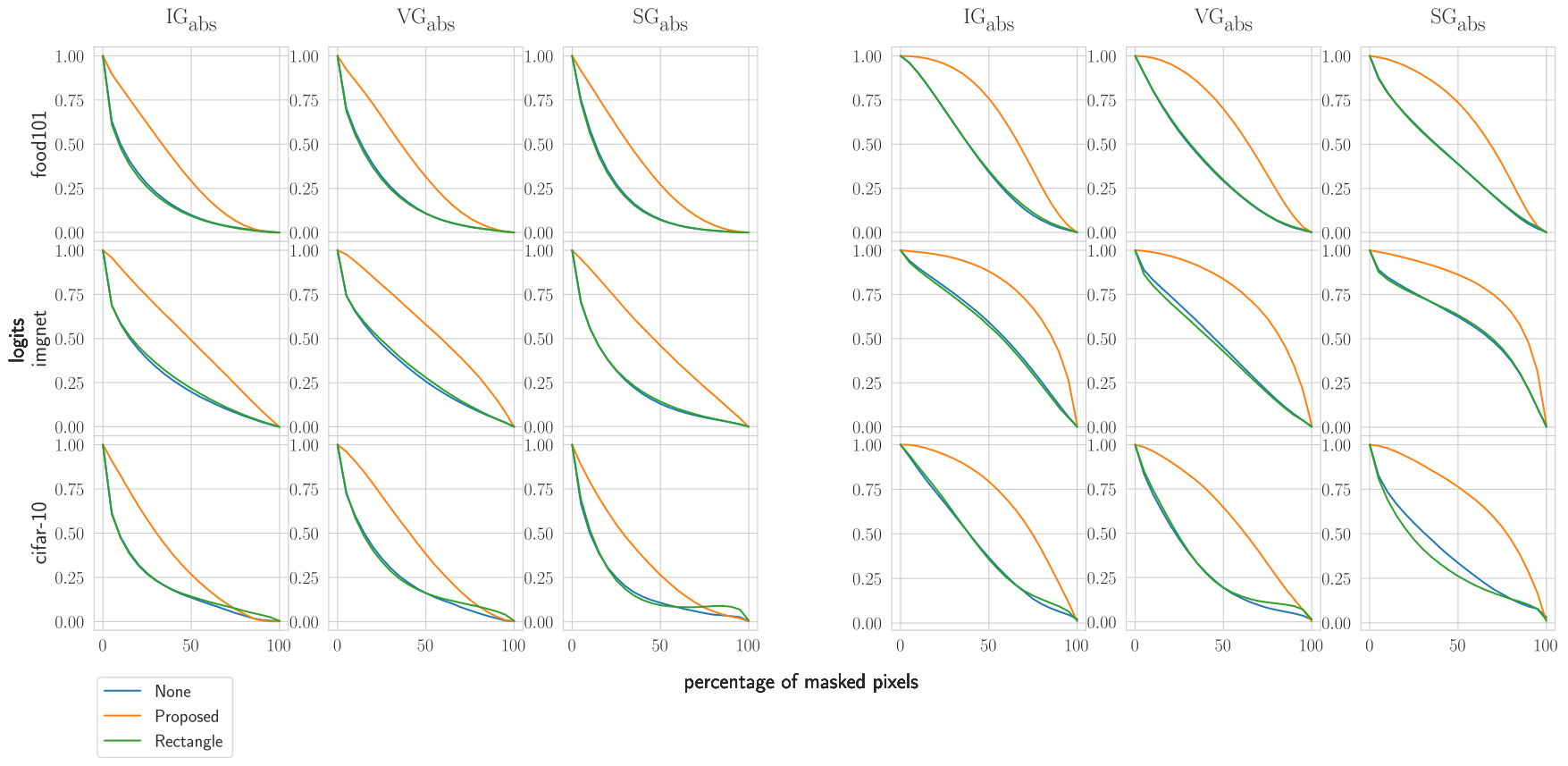}
\caption{Overview of perturbation curves for unsigned importance estimators for all three considered datasets. Sums of the absolute values of color channels are used. The change in normalized logits is plotted as a percentage of pixels that are masked according to the Most Important First (MIF; \emph{Left}) and the Least Important First (LIF; \emph{Right}). Notice that in comparison to the LIF perturbation curves from the signed estimators (\autoref{summary_signed}), the initial increase for LIF perturbation does not occur for the unsigned estimators.} 
\label{summary_unsigned}
\end{figure}

\begin{figure}[ht!]
\centering
\includegraphics[width=0.7\textwidth]{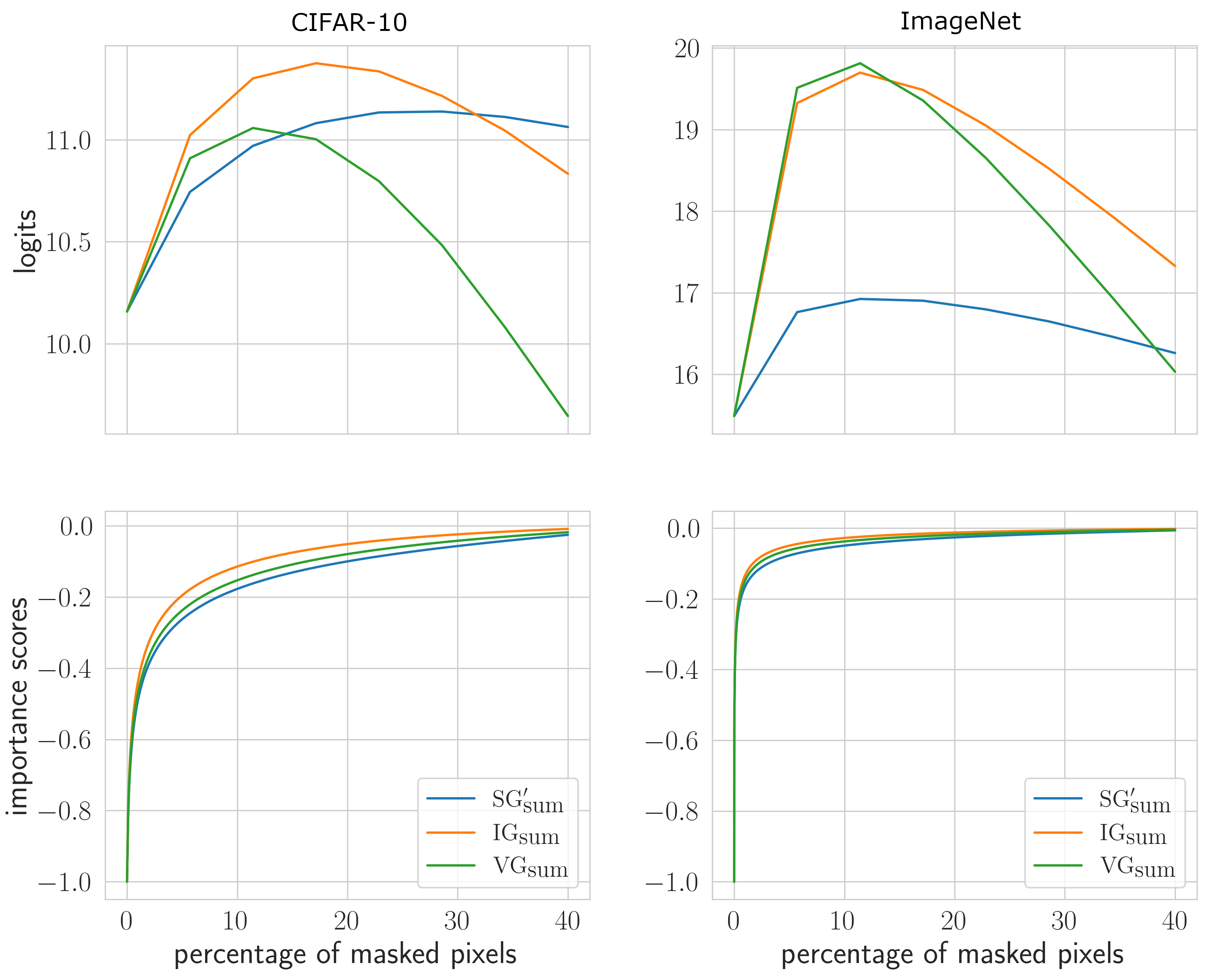}
\caption{\emph{Top row:} LIF perturbation curves averaged over CIFAR-10 and ImageNet test samples. \emph{Bottom row:} The curve has been obtained by flattening the heat map and plotting
the importance scores in LIF order, with the x-axis indicating their position in the ranking. At each point
on the x-axis, one can compare the importance scores of a group of pixels (y-axis labeled `importance scores') and the influence of their masking on the model output (y-axis labeled `logits').} 
\label{cifar_imgnet_compare}
\end{figure}

\clearpage
\section{Supplementary Tables}

\begin{table}[ht]
\centering
\caption{The fidelity of importance estimators $A$ (the area between LIF and MIF perturbation curves), measured on the ResNet-50 trained on ImageNet with $95\%$ confidence intervals. See the main text for the difference between the unsigned and signed importance estimators.}
\label{table:1}
\begin{tabular}{l c c c c c c c} 

 Aug. & Random & IG\ts{sum} & IG\ts{abs} & VG\ts{abs} & VG$'$\ts{sum} \\\midrule

 None & $0.0\pm 0.7$ & $15.9\pm 0.7$ & $29.2\pm 0.7$ & $14.6\pm 0.8$ & $6.2\pm 0.7$ \\ 
 Proposed & $0.0\pm 0.5$ & $\mathbf{61.8\pm 0.8}$ & $29.2\pm 0.6$ & $18.7\pm 0.6$ & $45.6\pm 0.9$ \\ \\

 Aug. & VG$'$\ts{abs}&SG\ts{abs} &  SG$'$\ts{sum} &SG$'$\ts{abs}& SQ-SG\ts{sum}\\\midrule

 None & $23.0\pm 0.8$ & $36.4\pm 0.7$ & $38.1\pm 0.9$ & $40.0\pm 0.7$& $\mathbf{42.7\pm 0.6}$ \\ 
 Proposed & $23.7\pm 0.6$ & $31.0\pm 0.6$ & $57.3\pm 0.8$ & $33.9\pm 0.6$ & $36.2\pm0.6$\\ 
\end{tabular}
\end{table}

\begin{table}[ht]
\centering
\caption{The fidelity of importance estimators $A$ (the area between LIF and MIF perturbation curves), measured on the ResNet-18 trained on CIFAR-10 with $95\%$ confidence intervals.}
\label{table:2}
\begin{tabular}{l c c c c c c} 

 Aug. & Random & IG\ts{sum} & IG\ts{abs}  & VG\ts{abs} & VG$'$\ts{sum}\\\midrule

 None & $0.0\pm 0.7$ & $14.0\pm 0.8$ & $21.9\pm 0.6$ & $5.6\pm 0.6$ & $8.0\pm 0.7$ \\ 
 Proposed & $0.0\pm 0.5$ & $59.6\pm 0.7$ & $34.9\pm 0.5$ & $16.9\pm 0.5$ & $47.0\pm 0.7$ \\ \\

 Aug. & VG$'$\ts{abs}  &SG\ts{abs} &  SG$'$\ts{sum} &SG$'$\ts{abs}& SQ-SG\ts{sum}\\\midrule

 None & $16.1\pm 0.6$ & $18.5\pm 0.6$ & $\mathbf{34.6\pm 0.8}$ & $27.0\pm 0.6$& $27.9\pm 0.6$\\ 
 Proposed & $25.8\pm 0.5$ & $35.4\pm 0.5$ & $\mathbf{80.5\pm 0.6}$ & $39.2\pm 0.5$ & $41.1\pm0.5$\\ 
\end{tabular}
\end{table}

\end{document}